\documentclass{article}

\usepackage[preprint]{neurips_data_2024}

\usepackage[colorinlistoftodos]{todonotes}

\usepackage[utf8]{inputenc} 
\usepackage[T1]{fontenc}    
\usepackage{hyperref}       
\usepackage{url}            
\usepackage{booktabs}       
\usepackage{amsfonts}       
\usepackage{nicefrac}       
\usepackage{microtype}      
\usepackage{xcolor}         

\usepackage{multirow}

\title{SeeFar: Satellite Agnostic Multi-Resolution Dataset for Geospatial Foundation Models}

\author{%
    James Lowman\textsuperscript{\thanks{ equal contribution} \hspace{1pt} \rm 1\rm 2},
    Kelly Liu Zheng\textsuperscript{\rm 1\rm 2},
    Roydon Fraser\textsuperscript{\rm 2},\\
    \textbf{Jesse Van Griensven The}\textsuperscript{\rm 2}, and
    \textbf{Mojtaba Valipour}\textsuperscript{$\ast$\rm 1\rm 2} \\
    \textsuperscript{\rm 1} Coastal Carbon, 
    \textsuperscript{\rm 2} University of Waterloo \\ 
    \texttt{\{mojtaba,kelly,james\}@coastalcarbon.ai}
}

\begin{document}

\maketitle

\begin{abstract}

SeeFar is an evolving collection of multi-resolution satellite images from public and commercial satellites. We specifically curated this dataset for training geospatial foundation models, unconstrained by satellite type. In recent years, advances in technology have made satellite imagery more accessible than ever. More earth-observing satellites have been launched in the last five years than in the previous fifty. Modern commercial satellites now offer up to 100 times the spatial resolution of public access satellites. However, the high cost and limited historical availability of commercial satellite imagery is a barrier to the training of foundational models, impacting what images can be used during inference. The SeeFar dataset represents a step towards training models that are satellite-agnostic by combining multi-resolution commercial and public access pre-processed images. This will enable users to utilize historical data alongside higher-resolution, more expensive satellite imagery, offering greater flexibility during inference. To achieve this, we describe a process for standardizing data from diverse satellite sources, normalizing different data formats, and aligning spectral bands to enhance interoperability. The SeeFar dataset includes images at a resolution of 384x384 pixels, spanning four spectral bands (Blue, Green, Red, and Near-Infrared) and expanding spatial resolutions (starting with 30, 10, 1.5, and 1.0 meters), all in cloud-optimized GeoTIFF format. It also provides consistent and comprehensive metadata to enhance data transparency and reliability. By aggregating data from multiple sources, SeeFar makes processed and consistent satellite data accessible to a wider range of users — from researchers to policymakers — fostering competition and innovation in satellite imagery analysis. The dataset is available at \url{coastalcarbon.ai/seefar}.

\end{abstract}

\section{Introduction}



Satellite imagery has revolutionized our understanding of the Earth, providing critical data for a wide range of applications, from environmental monitoring \cite{nerem2024satellite, chugg2022detecting, hu2021remote, young2017satellite} and urban planning \cite{meinel1997potential} to disaster response \cite{schumann2018assisting} and agricultural management \cite{steven1993satellite}. The ability to capture detailed images of the Earth’s surface from space has enabled scientists \cite{michener2011dataone}, policymakers \cite{lehmann2022essential}, and other professionals \cite{anderson2017earth} to make well-informed decisions using precise and up-to-date information provided by satellites.



\begin{figure}[ht!]
    \centering
    \resizebox{\textwidth}{!}{
    \includegraphics{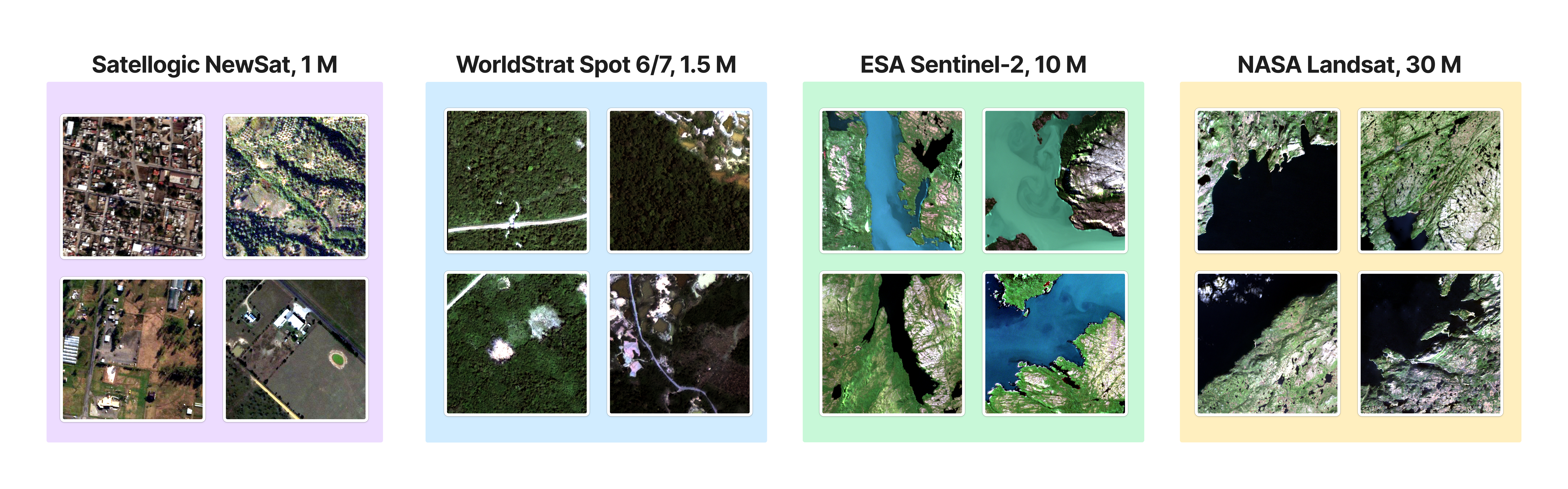}
    }
    \caption{SeeFar processed data examples from different satellite vendors.}
    \label{fig:examples}
\end{figure}

In recent years, technological advancements have dramatically increased the accessibility and quality of satellite imagery \cite{stamford2024remote, chetty2023satellite}. The past five years have seen more earth-observing satellites launched than in the previous fifty \cite{falle2023one}, with modern commercial satellites now offering up to 100 times the spatial resolution of traditional public access satellites \cite{stamford2024remote, brown1967synthetic}. These advancements have opened up new possibilities for high-resolution data analysis and application \cite{li2024automatic}.




Despite these advancements, significant challenges remain. The high cost of commercial satellite imagery (\$4 \footnote{\url{https://www.planet.com}} to \$30 \footnote{\url{https://up42.com/goingup/pleiades-neo}} per square kilometer) and its limited historical availability pose barriers to the training of foundational geospatial models. The development of such models is important because it will facilitate our understanding of the earth at scale. 

Satellite data is a unique modality \cite{rolf2024mission}. Not only is satellite data temporal, akin to video, but it is also spatially diverse with resolutions varying (as shown in Figure \ref{fig:examples}) from less than a meter to many kilometers \cite{zimmermann2024data}. Unlike conventional 3-channel RGB images in 8-bit depth, which have 256 possible values per pixel, satellite data can consist of multiple channels\footnote{\url{https://earth.esa.int/eogateway/missions/skysat}} \footnote{\url{https://modis.gsfc.nasa.gov/about/design.php}} captured in 12-bit depth, each pixel containing 4096 possible values. This significantly complicates the training of geospatial foundation models. Despite these challenges, there are a growing number of publications that aim to develop machine-learning models using satellite data, highlighting the importance of developing such models \cite{noman2024rethinking, li2024casformer, klemmer2023satclip, shenoy2024s4, cong2022satmae, xiong2024one, lacoste2024geo, xiong2024neural}.

Additionally, integrating and standardizing data from diverse sources, such as different types of satellites, remains a complex task. Diverse formats can lead to fragmented and inconsistent data, which hinders the development of robust, generalizable models that can effectively utilize data from multiple sources. This inconsistency directly impacts the performance and applicability of these models in real-world scenarios.







Without a dataset that seamlessly integrates historical data with high-resolution imagery, models are less likely to generalize well during inference since the majority of the accessible historical data is of low resolution. This limitation underscores the need for a dataset that bridges the gap between different satellite sources, enhancing the models’ robustness and flexibility.


\subsection{The problem} 
Current satellite datasets present several limitations that hinder their integration and analysis:

\begin{itemize}
    \item \textbf{Multi-Format Data}: Existing datasets \footnote{\url{https://huggingface.co/datasets/satellogic/EarthView}} come in various formats, making it challenging to effectively combine and analyze data from different sources.

    \item \textbf{Inconsistent Metadata}: Metadata inconsistencies across datasets impede accurate interpretation and comparison of satellite images, complicating tasks such as temporal analysis and multi-sensor fusion.

    \item \textbf{Temporal Scale Differences}: Variations in temporal resolution limit the ability to perform time-series analysis and monitor changes over consistent intervals.

    \item \textbf{Non-Uniform Spectral Bands}: Differing spectral bands make it difficult to standardize and normalize data for comparative studies or combined usage.

    \item \textbf{Variable Resolutions}: Differences in spatial resolution between datasets affect the accuracy and reliability of multi-resolution analysis.
\end{itemize}



A satellite-agnostic dataset like SeeFAR is crucial for addressing these challenges. SeeFAR aims to standardize data from diverse satellite sources, enhancing its applicability and integration into various analytical frameworks by normalizing and standardizing different satellite data formats, resolutions, and spectral bands. SeeFAR improves the interoperability of satellite imagery, facilitating more robust and scalable research and applications. It addresses temporal and spatial inconsistencies, enabling more accurate time-series analysis and multi-resolution studies. Furthermore, SeeFAR provides consistent and comprehensive metadata, enhancing data transparency and reliability, which is crucial for precise scientific analysis and decision-making. 

\subsection{Contributions}
We curated the SeeFar dataset specifically for training geospatial foundation models without resolution or satellite-type constraints. The following are the key contributions of our paper:

\begin{itemize}
    \item \textbf{Facilitate Training Foundation Models:} Foundation models, especially those trained on a single satellite source such as Landsat or Sentinel, struggle with adaptability to commercial imagery. To address this, we have assembled a dataset designed to train satellite-agnostic foundation models. By integrating multiple resolutions and harmonizing multispectral bands, our dataset aims to enhance the next generation of satellite image processing.
    
    \item \textbf{Standardization:} We developed a robust framework to standardize data from various satellite sources, ensuring a consistent dataset. This process normalizes different data formats, enhancing interoperability across diverse satellite imagery. 
    
    \item \textbf{Benchmarking:} The SeeFar dataset includes carefully curated training, validation, and test sets, enabling the fine-tuning and evaluation of geospatial models. These benchmarks facilitate the development of satellite-agnostic models, capable of handling imagery from both public and commercial satellites.

    \item \textbf{Multi-Resolution and Multi-Satellite Usability:} The dataset consists of an increasing number of 384x384 color images across four spectral bands: Blue, Green, Red, and Near-Infrared (NIR). These images are available at four different resolutions: 30 meters, 10 meters, 1.5 meters, and 1 meter. Images are curated from four different satellite platforms: Landsat 8 and 9, Sentinel 2, NewSat IV, and Spot 6 and 7. This multi-resolution and multi-satellite approach supports a wide range of applications, making the dataset suitable for various geospatial analysis tasks.
    
    \item \textbf{Metadata Augmentation:} SeeFar provides consistent and comprehensive metadata. In some cases, to ensure consistency we needed to augment the provider dataset. This metadata aids users and models in understanding the source, quality, and characteristics of the imagery, fostering more informed analyses.
\end{itemize}


The rest of the paper is organized as follows: we first present related work, then discuss our methodology, and then introduce the data sources along with the processes applied to each individual dataset. Finally, we address data licensing and summarize the paper with the conclusion and limitations.

\section{Related Work}


The field of satellite imagery and geospatial data has seen substantial advancements through several notable contributions. This section summarizes these contributions and highlights how our dataset addresses existing gaps and builds upon previous work.


Relevant to our work, the MuRA-T dataset \cite{deshmukh2023aligned} offers a sophisticated approach to aligning multi-satellite images, incorporating both Landsat and Sentinel datasets to enhance the SpaceNet-7 benchmark. While this dataset effectively tackles temporal and spatial alignment challenges, such big constraint will limit the scope of possible satellite sources, which can constrain its applicability for pre-training foundational models. Our dataset builds on this by relaxing the constraint, and therefore standardizing data from a broader range of satellites, including commercial ones, thus enhancing diversity and applicability.



\cite{deshmukh2023aligned}, \cite{francis2024major}, and \cite{cornebise2022open} have also presented frameworks and datasets that include data relevant to the multi-resolution task. However, these frameworks lack a consistent process for handling images from different satellite providers, leaving the integration and standardization to the end user. In contrast, we present a dataset and a framework designed to standardize data across different vendors, ensuring uniformity and ease of use for the end user.




The MultiSenGE dataset \cite{wenger2022multisenge} is a collection of 8157 patches covering a significant administrative region in Eastern France, including patches for Sentinel-2 L2A, Sentinel-1 GRD, and a regional LULC topographic database. While it provides detailed metadata for transparency, its regional focus limits broader applicability. Our dataset extends the utility of such detailed metadata by offering a global dataset, ensuring consistent and comprehensive metadata that enhances data transparency and reliability on a wider scale.


The EarthNet2021 dataset \cite{kim2023earthnet} is a collection of 32000 minicubes, each containing 30 frames, specifically curated for the task of earth surface forecasting aimed at forecasting the localized effects of severe weather conditions. 
However, the highest spatial resolution in this data is 20m which significantly hinders the use of this dataset for applications that need higher resolutions.

Additionally, the aligned multi-temporal multi-resolution dataset by \cite{spaceNet2024} provides a detailed framework for change detection, aligning high-resolution Planet images with Landsat and Sentinel data. While comprehensive, its focus on specific temporal alignments can limit flexibility. We enhance this approach by offering multi-resolution data across a wider range of satellites, enabling more versatile and scalable earth observation insights.

These contributions collectively illustrate the progress and ongoing efforts in satellite imagery analysis, laying a foundation for the further development and utilization of our dataset.

\section{Methodology}

\begin{figure}[ht!]
    \centering
    \resizebox{1.0\linewidth}{!}{
    \includegraphics{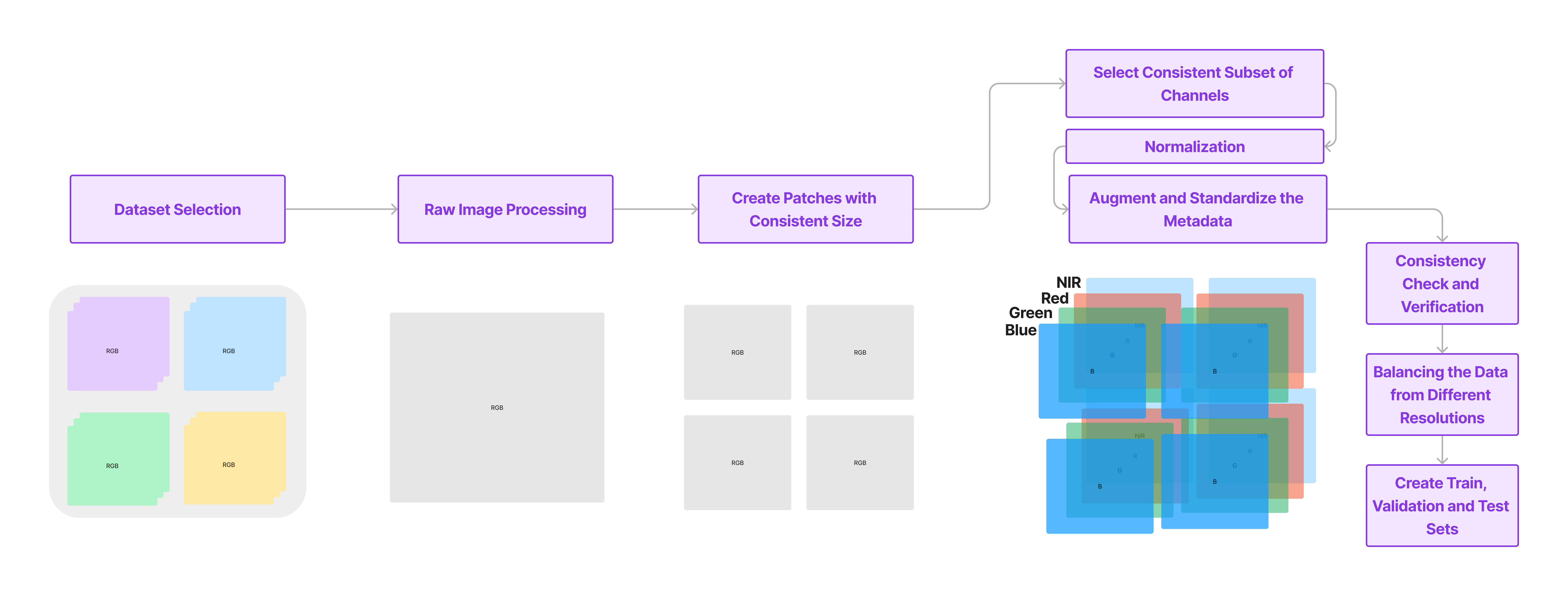}}
    \caption{SeeFar Methodology: an overview of the proposed method utilized to create a standard dataset.}
    \label{fig:method}
\end{figure}


Given a dataset of samples $x^{c\times d} \in X^{n\times c\times d}_S$ from a satellite S where n is the batch size, c is the number of spectral bands, and d is number of pixels, we want to create a dataset $D=\{\forall x: x \in \bigcup_{i=1}^N D_i\}$ with consistent examples across different datasets $D_1$, $D_2$, ..., $D_N$. Our proposed methodology consists of the following steps, as shown in Figure \ref{fig:method}: 

\paragraph{Dataset Selection:} Select a unique subset of commercial and public datasets that are publicly licensed, are rich in terms of content, alignment across datasets is possible, have unique resolutions, and their license allows for the data  to be shared or modified. Therefore, we chose Earthview \footnote{\href{https://huggingface.co/datasets/satellogic/EarthView}{\url{huggingface.co/datasets/satellogic/EarthView}}}, Worldstrat \cite{cornebise2022open}, Sentinel 2 \footnote{\href{https://www.esa.int/Applications/Observing_the_Earth/Copernicus/Sentinel-2}{\url{esa.int/Applications/Observing\_the\_Earth/Copernicus/Sentinel-2}}}, and Landsat \footnote{\href{https://landsat.gsfc.nasa.gov}{\url{landsat.gsfc.nasa.gov}}}. 

\paragraph{Raw Image Processing:} As each dataset provider has their own processing mechanism, raw files may not provide consistent information. Therefore, we must pre-process each data source differently. The details of each pre-processing mechanism can be found in the following section \ref{sec:data_sources}. 

\paragraph{Create Patches with Consistent Size:} We chose to use a consistent patch size to ensure that any model trained on SeeFar will focus on the semantic difference between data with different spatial resolutions and not the aspect ratio or other potential confounds that could bias the task. Following Earthview's choice of 384$\times$ 384 patch sizes, we generated the rest of the data to have the same height and width to ensure consistency. 


\paragraph{Select Consistent Subset of Channels:} We chose Blue, Green, Red, and NIR (data channels adhere to this order) as the selected spectral bands, as they are consistently available across the chosen datasets.

\paragraph{Normalization:} To ensure that the model will not simply learn to categorize different data sources from their brightness or even contrast, we used histogram matching to ensure the colour, brightness and contrast across different data samples are consistent.

\paragraph{Augment and Standardize the Metadata:} We augmented the metadata of each sample with the Ground Sample Distance (GSD), spectral wavelength bounds, and updated georeferencing. 

\paragraph{Cleaning, Consistency Check and Verification:} We also adapt automated mechanisms to double-check the consistency across different datasets and to ensure that the quality of the data is acceptable. This includes rejecting any sample with more than 5\% of pixels identified as 'no-data', and rejecting samples with more than 50\% of pixels identified as clouds. Cloud identification is accomplished by combining thresholding above 0.7 in both a normalized difference snow index (NDSI) and NIR band, following \cite{CHOI2004237}. 

\begin{equation}
    NDSI = \frac{Green - NIR}{Green + NIR}
\end{equation}

\paragraph{Balancing the Data from Different Resolutions:} We realize the importance of balancing the sample size from different data providers and, therefore, have curated the data with selective examples to avoid bias in the training phase.

\paragraph{Create Benchmarks:} At the end, we will divide the whole processed dataset into three sets before each release. One for training, the other one for validation, and the last as the test set.


\section{Dataset}

\subsection{Data Sources}
\label{sec:data_sources}


\paragraph{Satellogic NewSat IV}

The Satellogic NewSat IV dataset, sourced from Satellogic’s first open-sourced image set, “Earthview,” was selected for its high-resolution capabilities and compatibility with our multi-resolution approach. Each image in this dataset is 384x384 pixels, establishing the standard dimensions for all other datasets in our collection. This consistency in image size is crucial for uniform data processing and analysis across different sources.

Initially, each item in the dataset was in two separate arrays, one for RGB bands and another for NIR bands. To ensure data integrity and usability, we recombined these bands and reconstructed the metadata for each image into a new GeoTIFF file. The original arrays, down-sampled to 8-bit, were converted back to 16-bit to maintain uniformity with the other datasets despite the inherent loss of information due to the initial down-sampling.

We manually integrated custom metadata, including a Ground Sample Distance (GSD) of 1 meter, and detailed spectral band information specific to Satellogic’s ‘NewSat IV’ fleet instruments. This process of recombining spectral data and enhancing metadata ensures that the Satellogic NewSat IV dataset provides reliable and high-quality imagery, supporting comprehensive analysis and training of geospatial models.

\paragraph{WorldStrat Spot 6/7}

The WorldStrat Spot 6/7 dataset, sourced from the high-resolution portion of the “WorldStrat” \cite{cornebise2022open} collection, was chosen for its superior resolution and enhanced imagery quality. We utilized pansharpened images that have been upscaled to a 1.5-meter resolution by combining 1.5-meter panchromatic images with 6-meter multispectral data. This process results in new 1.5-meter multispectral images, significantly improving the detail and clarity of the imagery.

To ensure consistent patch sizes, we applied a sliding window technique to generate nine samples from each original 1054x1054 image. We also undertook a comprehensive metadata reconstruction process, extracting the original WorldStrat metadata, including the Coordinate Reference System (CRS) and transform details, from the raw data and reattaching it to the high-resolution images. This step ensures the integrity and accuracy of the spatial information associated with each image.

Furthermore, we added custom metadata, including a GSD of 1.5 meters and detailed spectral band information for Spot 6/7. This enhanced metadata provides users with critical information about the source and quality of the imagery, facilitating more informed analysis and application.

\paragraph{ESA Sentinel-2}

The ESA Sentinel-2 dataset, obtained from the “Planetary Computer” \footnote{\url{https://planetarycomputer.microsoft.com}} repository, was selected for its high-quality, freely available imagery and comprehensive temporal coverage. Each image in this dataset includes the blue (B02), green (B03), red (B04), and NIR (B08) bands, all at a native 10-meter GSD. We specifically use the “Level 2A” data, which is orthorectified to ensure accurate spatial representation. This orthorectification is consistent with all the data we use, ensuring uniformity across our datasets.

Processing Sentinel-2 data involves assembling the individual band GeoTIFFs into a single file containing all four bands. The general metadata from the blue band is copied into this new combined GeoTIFF to maintain consistency. We then segment the image into 384x384 pixel chunks, aligning with the dimensions used in other datasets for seamless integration and analysis. Chunks containing more than 5\% 'no-data' pixels are discarded to minimize 'no-data' areas.

Additionally, we apply custom metadata to each segmented image, enhancing the dataset’s utility. This metadata includes critical information about the image source, spectral bands, and resolution, facilitating more informed and accurate analyses.

\paragraph{NASA Landsat}

The NASA Landsat dataset, sourced from the “Planetary Computer” repository, was chosen for its extensive historical coverage and reliable, freely available imagery. Each image in this dataset includes the blue, green, red, and NIR bands, all at a native 30-meter GSD. We utilize the “Level 2” data, which is orthorectified to ensure accurate spatial representation, maintaining consistency across our datasets.

Processing Landsat data involves assembling the individual band GeoTIFFs into a single file containing all four bands. The general metadata from the blue band is copied into this new combined GeoTIFF to ensure uniformity. We then segment the image into 384x384 pixel chunks, removing any chunk with more than 5\% 'no-data' pixels, aligning with the dimensions and consistency used in other datasets for seamless integration and analysis.

Additionally, we apply custom metadata to each segmented image, enhancing the dataset’s utility. This metadata includes GSD, spectral bands, and resolution, facilitating more informed and accurate analyses.

\begin{table*}[h!]
    \centering
    \caption{The table presents the spectral information for different satellite data sources. For NewSat IV, the details are reported from \cite{agliozzo2023orbit}, for Spot 6/7 from \cite{app10051881}, and for Sentinel-2 and Landsat 8/9 from \cite{cerasoli2018estimating}. All data in this table is encoded in the metadata for each image in SeeFar.}
    \begin{tabular}{|cccc|}
        \hline
        \textbf{Satellite Source} & \textbf{Band} & \textbf{Start (nm)} & \textbf{End (nm)} \\
        \hline
        \multirow{4}{*}{Satellogic NewSat IV} & Blue & 450 & 510 \\
         & Green & 510 & 580 \\
         & Red & 590 & 690 \\
         & NIR & 750 & 900 \\
        \hline
        \multirow{4}{*}{Spot 6/7} & Blue & 450 & 520 \\
         & Green & 530 & 590 \\
         & Red & 625 & 695 \\
         & NIR & 760 & 890 \\
        \hline
        \multirow{4}{*}{ESA Sentinel-2} & Blue & 458 & 523 \\
         & Green & 543 & 578 \\
         & Red & 650 & 680 \\
         & NIR & 785 & 899 \\
        \hline
        \multirow{4}{*}{NASA Landsat 8/9} & Blue & 452 & 512 \\
         & Green & 533 & 590 \\
         & Red & 636 & 673 \\
         & NIR & 851 & 879 \\
        \hline
    \end{tabular}
    \label{tab:spectral_info}
\end{table*}






\subsection{Licenses}
\label{sec:licenses}

The datasets utilized in this research were obtained under a variety of licenses, each with specific usage terms and restrictions. Satellogic data from the Earthview are provided under the Creative Commons Attribution 4.0 (CC-BY 4.0) license. This allows for free use, sharing, and adaptation, provided appropriate credit is given and any changes are indicated.

WorldStrat's high-resolution Airbus imagery is available under the Creative Commons Attribution-Non Commercial 4.0 (CC-BY-NC 4.0) license, which shares the same requirements as CC-BY 4.0 but restricts use to non-commercial purposes.


Sentinel satellite data is governed by the Creative Commons Attribution-Share Alike 3.0 IGO (CC-BY-SA 3.0 IGO) license, which requires attribution, a link to the license, and an indication of any changes. If building upon this material, the newly developed material must be distributed under the same license.

Finally, USGS Landsat data is available under the USGS Landsat License, which places no restrictions on usage or redistribution but requests a statement of the data source when citing, copying, or reprinting.

\section{Conclusion}

In conclusion, SeeFar represents a significant step forward in the development of satellite-agnostic geospatial models and the integration of multi-resolution satellite imagery. By harmonizing data from diverse satellite sources and standardizing various formats, spectral bands, and resolutions, SeeFar provides a versatile and comprehensive dataset that addresses a current limitation in satellite data integration and analysis.

As SeeFar continues to evolve, ongoing updates and expansions will aim to incorporate a broader range of resolutions and further improve data quality and consistency. This commitment to continuous improvement ensures that SeeFar will remain a valuable resource for researchers, policymakers, and industry professionals, fostering innovation and advancing the field of satellite imagery analysis.

Overall, SeeFar provides widespread access to high-quality satellite data, enabling more informed decision-making and supporting a broad spectrum of geospatial research and practical applications. The ongoing development and refinement of this dataset will significantly contribute to meeting the growing demands of scientific research and addressing real-world challenges in the geospatial domain.

\section{Limitation}
\label{sec:limitation}

SeeFar is an evolving dataset, and as of now, is limited to a subset of resolutions available from different satellite vendors. While the dataset integrates multi-resolution images from various sources, it does not encompass the full range of resolutions offered by all satellite providers. This limitation may impact the granularity and comprehensiveness of analyses performed using SeeFar, particularly for applications requiring extremely high-resolution imagery or specialized spectral bands not currently included in the dataset.

Moreover, as the dataset grows and evolves, it may become necessary to periodically update and reprocess existing data to incorporate new standards, corrections, and improvements. Users relying on the dataset for long-term projects should be prepared for potential changes and updates that could affect the consistency and continuity of the data over time.

\bibliography{custom}
\bibliographystyle{plainnat}






\newpage
\section*{Checklist}


\begin{enumerate}

\item For all authors...
\begin{enumerate}
  \item Do the main claims made in the abstract and introduction accurately reflect the paper's contributions and scope?
    \answerYes{}
  \item Did you describe the limitations of your work?
    \answerYes{Section~\ref{sec:limitation}}
  \item Did you discuss any potential negative societal impacts of your work?
    \answerNA{}
  \item Have you read the ethics review guidelines and ensured that your paper conforms to them?
    \answerYes{}
\end{enumerate}

\item If you are including theoretical results...
\begin{enumerate}
  \item Did you state the full set of assumptions of all theoretical results?
    \answerNA{}
   \item Did you include complete proofs of all theoretical results?
    \answerNA{}
\end{enumerate}

\item If you ran experiments (e.g. for benchmarks)...
\begin{enumerate}
  \item Did you include the code, data, and instructions needed to reproduce the main experimental results (either in the supplemental material or as a URL)?
    \answerNA{}
  \item Did you specify all the training details (e.g., data splits, hyperparameters, how they were chosen)?
    \answerNA{}
	\item Did you report error bars (e.g., with respect to the random seed after running experiments multiple times)?
    \answerNA{}
	\item Did you include the total amount of compute and the type of resources used (e.g., type of GPUs, internal cluster, or cloud provider)?
    \answerNA{}
\end{enumerate}

\item If you are using existing assets (e.g., code, data, models) or curating/releasing new assets...
\begin{enumerate}
  \item If your work uses existing assets, did you cite the creators?
    \answerYes{}
  \item Did you mention the license of the assets?
    \answerYes{Section~\ref{sec:licenses}}
  \item Did you include any new assets either in the supplemental material or as a URL?
    \answerYes{Link to the dataset is in the abstract}
  \item Did you discuss whether and how consent was obtained from people whose data you're using/curating?
    \answerNA{All datasets are publicly accessible, we checked their licenses with legal advisors.}
  \item Did you discuss whether the data you are using/curating contains personally identifiable information or offensive content?
    \answerNA{All data are publicly accessible.}
\end{enumerate}

\item If you used crowdsourcing or conducted research with human subjects...
\begin{enumerate}
  \item Did you include the full text of instructions given to participants and screenshots, if applicable?
    \answerNA{}
  \item Did you describe any potential participant risks, with links to Institutional Review Board (IRB) approvals, if applicable?
    \answerNA{}
  \item Did you include the estimated hourly wage paid to participants and the total amount spent on participant compensation?
    \answerNA{}
\end{enumerate}

\end{enumerate}


\end{document}